\documentclass[a4paper]{article}

\usepackage[english]{babel}
\usepackage[utf8x]{inputenc}
\usepackage[T1]{fontenc}

\usepackage[a4paper,top=3cm,bottom=2cm,left=3cm,right=3cm,marginparwidth=1.75cm]{geometry}

\usepackage{amsmath}
\usepackage{graphicx}
\usepackage[colorinlistoftodos]{todonotes}
\usepackage[colorlinks=true, allcolors=blue]{hyperref}
\usepackage{hyperref}
\usepackage{url}
\usepackage{amsmath}
\usepackage{amsfonts}
\usepackage{amssymb}
\usepackage{graphicx}
\usepackage{graphics}
\usepackage{hyperref}
\usepackage{mathrsfs}
\usepackage{multirow}
\usepackage{float}
\usepackage{amsfonts}       
\usepackage{nicefrac}       
\usepackage{microtype}      
\usepackage{amsmath}
\usepackage{amsfonts}
\usepackage{amssymb}
\usepackage{graphicx}
\usepackage{subfigure}
\usepackage{graphics}
\usepackage{multirow}
\usepackage{float}
\usepackage{bm}
\usepackage{mathrsfs}
\usepackage{color}
\usepackage{algorithm}
\usepackage{algorithmic}
\usepackage{mathrsfs}
\usepackage{cite}
\usepackage{diagbox}

\title{A Simple Yet Efficient Rank One Update for Covariance Matrix Adaptation}
\author{Zhenhua Li,
	and~Qingfu Zhang\\
	Department of Computer Science\\
	 City University of Hong Kong, Hong Kong\\
		e-mail: zhenhua.li@my.cityu.edu.hk.}
\date{}
\begin{document}
	\maketitle

\begin{abstract}
In this paper, we propose an efficient approximated rank one update for covariance matrix adaptation evolution strategy (CMA-ES). It makes use of two evolution paths as simple as that of CMA-ES, while avoiding the computational matrix decomposition. We analyze the algorithms' properties and behaviors. We experimentally study the proposed algorithm's performances. It generally outperforms or performs competitively to the Cholesky CMA-ES. 
\end{abstract}

\section{Introduction}
Evolution strategies (ES) are one of the main streams of evolutionary algorithms~\cite{Beyer02} for continuous black-box optimization~\cite{Kern04a}\cite{Posik2010}. Typically, ES evolves a Gaussian distribution~\cite{Beyer02}. It iteratively samples one or more new candidate solutions, and adapts the distribution parameters for producing more promising solutions. The covariance matrix adaptation evolution strategy (CMA-ES) is one of the most successful evolution strategies. It adapts a full covariance matrix, which achieves invariant under general linear transformation in the search space~\cite{Hansen08}\cite{Hansen14}. 

The covariance matrix is adapted to increase the probability of reproducing the successful search directions in the previous generations. In practice, it is updated by the rank-$\mu$ update with maximum likelihood estimation of successful solutions and the rank one update with the evolution paths, which involves the historical search information. A main limitation of CMA-ES is that it has to conduct matrix decomposition to generate new solutions~\cite{Hansen01}\cite{Hansen03}\cite{Beyer06}, which is computationally expensive. This prohibits its applications to problems with over hundreds of variables. In practice, the matrix decomposition can be done every $O(n)$ generations to alleviate the computational overload, during which the outdated matrices are used to generate new solutions. This is undesirable as it does not make use of the current search information, and may result in some performance loss~\cite{Suttorp09}.

To alleviate the issue of computational overload, Cholesky CMA-ES directly adapts the Cholesky factor instead of the covariance matrix~\cite{Igel2006}. It considers a rank one perturbation of the covariance matrix, and obtains a rank one update for the Cholesky factor. This achieves complexity $O(n^2)$. To use the evolution path in Cholesky CMA-ES, it has to maintain and update the inverse Cholesky factor accordingly~\cite{Suttorp09}, which is much complicated than the rank one update for the covariance matrix. It becomes even more complicated when restricting the Cholesky factor to be upper triangular~\cite{Krause15}. 


In this paper, we propose to construct a new evolution path to approximate the inverse vector of the evolution path $\mathbf{A}^{-1}\mathbf{p}$. Then both evolution paths contributes an efficient rank one update for the mutation matrix $\mathbf{A}$ that satisfies $\mathbf{AA}^T =\mathbf{C}$. It avoids maintaining the inverse Cholesky factor, and reduces the computational complexity in the update procedure of Cholesky CMA-ES to a half. The obtained algorithm is as simple as CMA-ES with pure rank one update~\cite{Hansen01}.

We analysis the approximations in the proposed algorithm, and validate the approximations experimentally. We experimentally study the algorithm performances. It shows that the proposed methods outperform or performs competitively to Cholesky CMA-ES.


\section{Background}

In this section, we consider the black-box optimization in the continuous domain
\begin{equation}
	\min_{\mathbf{x}\in\mathbb{R}^n } f(\mathbf{x}).
\end{equation}
In black-box optimization, we have no gradient information of $f$ available, and the only information accessible is the objective value for a given point $\mathbf{x} \in \mathbb{R}^n$.
\subsection{Gaussian Mutations in Evolution Strategies}
In a canonical ES, a new candidate solution is generated (at generation $t$) by
\begin{equation}\label{sample}
\mathbf{x}  = \mathbf{m}_t + \sigma_{t} \mathbf{y},
\end{equation}
where $\mathbf{m}_t$ is the current distribution mean, determining the location of the mutation distribution over the search space, and $\sigma_t$ is the step size which determines the global variance of the underlying distribution. The mutation $\mathbf{y} \in \mathbb{R}^n$ is a $n$-dimensional random vector distributed according to the Gaussian mutation distribution
\begin{equation}
\mathbf{y}  \sim \mathcal{N}(\mathbf{0}, \mathbf{C}_{t}).
\end{equation} 
The approaches that evolving arbitrary Gaussian mutation distributions introduce complete invariance with respect to translation and rotation in the search space~\cite{Hansen11}. 

To draw a mutation from the Gaussian distribution $\mathcal{N}(\mathbf{0}, \mathbf{C}_{t})$, it is necessary to decompose the covariance matrix. In CMA-ES, the eigen-decomposition is conducted as $\mathbf{C}_{t} =\mathbf{B} \mathbf{D}^2\mathbf{B}^T$, where $\mathbf{B}$ is orthogonal matrix with eigenvectors as columns, and $\mathbf{D}$ is a diagonal matrix with the square root of eigenvalues as the entries. A new solution is generated by
\begin{equation}\label{CMAsample}
\mathbf{x}  = \mathbf{m}_t + \sigma_{t} \mathbf{BDz},  ~ \mathbf{z} \sim \mathcal{N}(\mathbf{0},\mathbf{I}).
\end{equation}
An alternative is to use the Cholesky decomposition~\cite{Beyer06}. The computational complexity of the matrix decomposition is $O(n^3)$ in general, which dominates the computational complexity of CMA-ES.


\subsection{The Parameter Dependencies}

Evolving a full covariance matrix renders the invariance under generalized linear transformations in the search space~\cite{Hansen08}\cite{Hansen14}. The covariance matrix or its factorizations actually present a transformation in the search space, which transforms the objective function to be more separable and well-scaled~\cite{Hansen08}. The transformed objective function can be solved more efficiently. Hence, it is of crucial importance to learn the local transformation.

\subsubsection{Covariance Matrix Adaptation}
CMA-ES evolves a full covariance matrix to represent the parameter dependencies. It updates the covariance matrix by the maximum likelihood estimation of the successful mutations. The original CMA-ES~\cite{Hansen01} with pure rank one update is 
\begin{equation}\label{rank1}
\mathbf{C}_{t+1}= (1-c_1)\mathbf{C}_{t}+c_1 \mathbf{p}_{t+1}\mathbf{p} _{t+1}^T.
\end{equation}
The evolution path $\mathbf{p}$ accumulates the movements of the distribution mean in consecutive generations and represents one of the most promising search directions~\cite{Li2016}. 

To make better use of the current population, it also uses the search directions of the selected solutions. Thus, the update for covariance matrix is given by
\begin{equation}\label{rankmu}
\mathbf{C}_{t+1}= (1-c_1-c_{\mu})\mathbf{C}_{t}+c_1 \mathbf{p}_{t+1}\mathbf{p} _{t+1}^T+ c_{\mu}\sum_{i=1}^{\mu}w_i \mathbf{y}_{i:\lambda} \mathbf{y} _{i:\lambda}^T,
\end{equation}
where $c_1, c_{\mu}$ are constant changing rates, and $\mathbf{y}_{i:\lambda}$ denotes the mutation corresponding to the $i$-th best solution of the current population. The rank-$\mu$ update increases the variance of successful search directions by a weighted maximum likelihood estimation, and increases the probability of reproducing such directions.


\subsubsection{Cholesky Covariance Matrix Adaptation}

To avoid the costly matrix decomposition, Cholesky CMA-ES directly adapts the Cholesky factor of the covariance matrix $\mathbf{C}_t = \mathbf{A}_t\mathbf{A}_t^T$. From the rank one update for the covariance matrix, it obtains the rank one Cholesky update~\cite{Igel2006}. 

To make use of the evolution path, it has to additionally maintain and adapt the inverse Cholesky factor $\mathbf{A}_t^{inv}$ to avoid the expensive computation of $\mathbf{A}_t^{-1}\mathbf{p}_{t+1}$. This makes the adaptation scheme complicated~\cite{Suttorp09}. The inverse vector of the evolution path is given by 
\begin{equation}\label{uvector}
\mathbf{u}_{t+1}= \mathbf{A}_{t}^{inv} \mathbf{p}_{t+1},
\end{equation}
where $\mathbf{A}_{t}^{inv}$ approximates $\mathbf{A}_t^{-1}$. Then both matrices are updated by 
\begin{align}
\mathbf{A}_{t+1} &= \sqrt{1-c_1} \mathbf{A}_t + \frac{\sqrt{1-c_1}}{\| \mathbf{u}_{t+1}\|^2} \left(\sqrt{1+\frac{c_1}{1-c_1}\|\mathbf{u}_{t+1}\|^2}-1 \right) \mathbf{p}_{t+1} \mathbf{u}_{t+1}^T,\\
\mathbf{A}_{t+1}^{inv} &=\frac{1}{\sqrt{1-c_1}} \mathbf{A}_t^{inv} + \frac{1}{\sqrt{1-c_1 }\| \mathbf{u}_{t+1}\|^2 }\left(\frac{1}{\sqrt{1+\frac{c_1}{1-c_1} \|\mathbf{u}_{t+1}\|^2}}-1 \right) \mathbf{u}_{t+1} \left(\mathbf{u}_{t+1}^T \mathbf{A}_t^{inv}\right).
\end{align}
This update scheme reduces the internal complexity to $O(n^2)$. However, it is much more complicated than the rank one update of CMA-ES~\eqref{rankmu}.

\section{Proposed Algorithm}

We propose an efficient update scheme for the matrix $\mathbf{A}$ that satisfies $\mathbf{C}\approx \mathbf{AA}^T $, where $\mathbf{A}$ can be referred to mutation matrix. The algorithm maintains and evolves the following parameters at generation $t$:
\begin{itemize}
	\item $\mathbf{m}_t$: the distribution mean which determines the location of the sampling region.
	\item $\mathbf{A}_t$: the non-singular mutation matrix, determining the shape of the sampling distribution.
	\item $\sigma_t$: the step size which determines the global variance of the sampling region.
	\item $\mathbf{p}_t, \mathbf{v}_t$: the evolution paths used to update the mutation matrix.
	\item $\mathbf{s}_t$: the evolution path used to update the step size. 
\end{itemize}
For convenience, we denote the evolution paths as the $\mathbf{p}$-path, the $\mathbf{v}$-path, and the  $\mathbf{s}$-path, respectively. At the beginning, these distribution parameters are initialized to be $\mathbf{m}_t\in \mathbb{R}^n, \sigma >0,\mathbf{A}_{0} =\mathbf{I} $, and the evolution paths $\mathbf{0}$. 

\subsection{Main Procedures}
In the following, we present the major procedures of the algorithm.
\subsubsection{Sampling New Solutions}
At each generation after initialization, a number of $\lambda$ solutions are generated from the current distribution. A new solution $\mathbf{x}_i, ~i=1,\cdots,\lambda$ is obtained by
\begin{equation}\label{Cholsample}
\mathbf{x}_i = \mathbf{m}_t+\sigma_t \mathbf{A}_t \mathbf{z}_i \sim \mathcal{N}(\mathbf{m}_t, \sigma_t^2 \mathbf{A}_t \mathbf{A}_t^T ),
\end{equation}
where $\mathbf{z}_i \sim \mathcal{N}(\mathbf{0}, \mathbf{I})$. The vector $\mathbf{z}_{i}$ is transformed by the matrix $\mathbf{A}_t $ to favor the successful search directions in previous generations. 

\subsubsection{Evaluation and Selection}
The objective values of the newly sampled solutions are computed to evaluate their qualities. Then they are sorted according to the objective values 
\begin{equation}
f(\mathbf{x}_{1:\lambda})\leq  \dots \leq f(\mathbf{x}_{\lambda:\lambda}),
\end{equation}
where $\mathbf{x}_{i:\lambda}$ denotes the $i$-th best solution of the current population. For truncation selection, the best $\mu \leq \frac{\lambda}{2}$ ones are selected to update the distribution parameters. The distribution mean is updated as
\begin{equation}
\mathbf{m}_{t+1} =\sum_{i=1}^{\mu}w_i\mathbf{x}_{i:\lambda},
\end{equation}
where $w_i$ can be set to $w_i = 1/\mu$ for $i=1,\dots, \mu$. A better choice is given by 
\begin{equation}
w_i =\frac{\ln(\mu+1)-\ln i }{\mu \ln(\mu+1)-\sum_{j=1}^{\mu}\ln j }.
\end{equation}
It is designed $w_1\geq \dots \geq w_{\mu}$ to favor the top ranked solutions, and $\sum_{i=1}^{\mu} w_i=1$. 
\begin{figure}[!htb]
	\centering
	\includegraphics[width=0.5\linewidth]{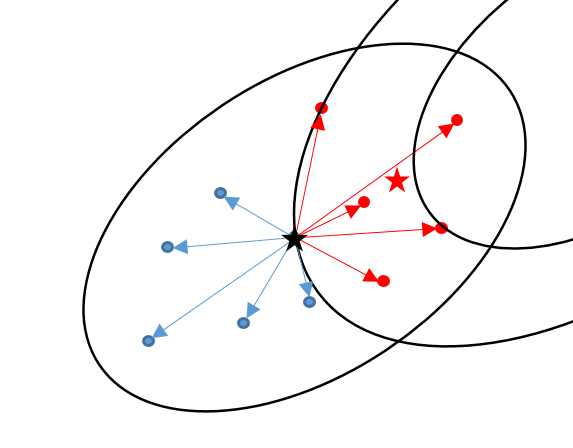}
	\caption{Selection and intermediate multi-recombination.}
	\label{sampling}
\end{figure}
\subsubsection{Evolution Paths}
The matrix $\mathbf{A}$ is updated by evolution paths. For convenience, we denote
\begin{equation}
\mathbf{z}_w =  \sum_{i=1}^{\mu}w_i\mathbf{z}_{i:\lambda}, \quad \mathbf{y}_w =  \sum_{i=1}^{\mu}w_i \left(\frac{\mathbf{x}_{i:\lambda}-\mathbf{m}_t}{\sigma_t}\right),
\end{equation}
where $\mathbf{z}_{i:\lambda}$ corresponds to the $i$-th best solution $\mathbf{x}_{i:\lambda}$. Obviously, we have $\mathbf{y}_w = \mathbf{A}_t\mathbf{z}_w$.

We construct and update the evolution paths as
\begin{align}
\mathbf{p}_{t+1} &= (1-c)\mathbf{p}_{t} + \sqrt{c(2-c)}  \sqrt{\mu_{\text{eff}}}\mathbf{y}_w, \label{ppath}\\
\mathbf{v}_{t+1} &= (1-c)\mathbf{v}_{t} + \sqrt{c(2-c)}  \sqrt{\mu_{\text{eff}}}\mathbf{z}_w. \label{vpath}
\end{align}
The involved parameters are given in the following.
\begin{itemize}
	\item The factor $\mu_{\text{eff}} = 1/\sum_{i=1}^{\mu}w_i^2$ normalizes the update direction such that under random selection, we have
	\begin{equation}
	\sqrt{\mu_{\text{eff}}}\mathbf{y}_w \sim\mathcal{N}(\mathbf{0}, \mathbf{A}_t\mathbf{A}_t^T ), ~\sqrt{\mu_{\text{eff}}}\mathbf{z}_w \sim\mathcal{N}(\mathbf{0}, \mathbf{I}) .
	\end{equation}
	\item The $\sqrt{c(2-c)}$ is determined by
	\begin{equation}
	(1-c)^2  + (\sqrt{c(2-c)})^2 = 1.
	\end{equation}
	This indicates that, given $\mathbf{p}_t \sim\mathcal{N}(\mathbf{0}, \mathbf{A}_t\mathbf{A}_t^T )$, we have $\mathbf{p}_{t+1} \sim\mathcal{N}(\mathbf{0}, \mathbf{A}_t\mathbf{A}_t^T )$ under the random selection. The similar condition holds for the $\mathbf{v}$-path that given $\mathbf{v}_t \sim\mathcal{N}(\mathbf{0}, \mathbf{I}) $, we have $\mathbf{v}_{t+1} \sim\mathcal{N}(\mathbf{0}, \mathbf{I}) $.
	\item The constant changing rate $c$ determines the number of generations involved in the cumulation of the evolution paths. Generally, the changing rate $c$ is set to $c \propto 1/n$ such that $c^{-1} \propto n$, which is the same order to the number of free parameters of $\mathbf{p}$ and $\mathbf{v}$.
\end{itemize}

The $\mathbf{p}$-path accumulates average successful search directions in consecutive generations. Averaging over generations, opposite search directions are canceled out and consistent search directions are accumulated. Hence, the $\mathbf{p}$-path represents the most promising search direction~\cite{Li2016}. 

Further, for $c=1$, no cumulations take place, and the equations~\eqref{ppath} and~\eqref{vpath} gives that $\mathbf{v}_{t+1} = \mathbf{A}_t^{-1} \mathbf{p}_{t+1}$. For $c \in (0,1)$, the cumulations take place in both evolution paths, and the $\mathbf{v}$-path accumulates $\mathbf{z}_w = \mathbf{A}_t^{-1}\mathbf{y}_w$ at each generation. Both $\mathbf{p}_{0}$ and $\mathbf{v}_{0}$ are initialized to zero, and the deviations of the $\mathbf{v}$-path from the $\mathbf{p}$-path in previous generations gradually damped by the factor $(1-c)$. Therefore, for $c \in (0,1)$, we have the following approximation in general
\begin{equation}
\mathbf{v}_{t+1} \approx \mathbf{A}_t^{-1} \mathbf{p}_{t+1}. 
\end{equation}

\subsubsection{Adaptation of the Mutation Matrix}
We use the $\mathbf{v}$-path to replace the inverse vector~$\mathbf{u}_{t+1}$, and update the mutation matrix as

\begin{equation}\label{HES}
\mathbf{A}_{t+1} = (1-\frac{c_1}{2}) \mathbf{A}_t + \frac{c_1}{2}\mathbf{p}_{t+1} \mathbf{v}_{t+1}^T.
\end{equation}
where $c_1$ is constant changing rate for the mutation matrix. Note that $\mathbf{p}_{t+1}\mathbf{v}_{t+1}^T$ is a $n\times n $ matrix with rank one. We denote this update scheme by mutation matrix adaptation evolution strategy (MMA-ES).

This rank one update can be obtained by analyzing the coefficients of the rank one update of the efficient Cholesky CMA-ES (eCMA-ES)~\cite{zhenhua17a}. In eCMA-ES, it updates the matrix $\mathbf{A}$ by two evolution paths as
\begin{equation}\label{eCMA}
\mathbf{A}_{t+1} = \sqrt{1-c_1}  \mathbf{A}_t + \frac{\sqrt{1-c_1}}{\| \mathbf{v}_{t+1}\|^2} \left(\sqrt{1+\frac{c_1}{1-c_1}\|\mathbf{v}_{t+1} \|^2}-1 \right) \mathbf{p}_{t+1}  \mathbf{v}_{t+1}^T.
\end{equation}
Substituting the approximation $\mathbf{v}_{t+1} \approx \mathbf{A}_t ^{-1}\mathbf{p}_{t+1}$, it is easy to verify that
\begin{equation*}
\mathbf{A}_{t+1}\mathbf{A}_{t+1}^T \approx (1-c_1)\mathbf{A}_t\mathbf{A}_t^T + c_1\mathbf{p}_{t+1}  \mathbf{p}_{t+1}^T.
\end{equation*}
It is the rank one update of the covariance matrix in CMA-ES.


The changing rate $c_1$ is set to $c_1 = 2 / (n + \sqrt{2})^2$ in the Cholesky CMA-ES~\cite{Suttorp09}. It gives that $c_1 \ll 1$ for even moderate $n$, which gives the approximation
\begin{equation*}
\sqrt{1 - c_1} \approx 1 - \frac{c_1}{2}.
\end{equation*}
Further, as the $\mathbf{v}$-path is normally distributed $\mathbf{v} \sim \mathcal{N}(\mathbf{0}, \mathbf{I})$ under random selection, its norm is distributed according to the Chi distribution $\| \mathbf{v} \| \sim \chi (n)$ and $\mathbb{E} \big[\| \mathbf{v} \|\big] = \sqrt{n}$. Hence, we generally have $\frac{c_1}{(1 - c_1)} \| \mathbf{v} \|^2  \ll 1$ for $\mathbf{v} \sim \mathcal{N}(\mathbf{0},\mathbf{I})$. Denote the coefficient by $b$, we obtain the following approximation
\begin{align*}
b & =\frac{\sqrt{1-c_1}}{\| \mathbf{v}_{t+1}\|^2} \left(\sqrt{1+\frac{c_1}{1-c_1}\|\mathbf{v}_{t+1} \|^2}-1 \right)\\
& \approx \frac{\sqrt{1 - c_1}}{\| \mathbf{v}_{t+1} \|^2} \cdot \frac{c_1}{2 (1-c_1)} \| \mathbf{v}_{t+1} \|^2\\
& = \frac{c_1}{2 \sqrt{1 - c_1}}\\
& \approx \frac{c_1}{2}.
\end{align*}
The approximations are due to $\sqrt{1+x} \approx 1+\frac{x}{2}$ for $ |x |\ll 1$. 

From the above approximations, we obtain the approximated rank one update for the matrix $\mathbf{A}$ given by equation~\eqref{HES}. The matrix $\mathbf{A}$ satisfies $\mathbf{C}\approx \mathbf{AA}^T$. Therefore, MMA-ES is a great simplification of the Cholesky CMA-ES.
\begin{figure}[!htb]
	\centering
	\includegraphics[width=0.75\linewidth]{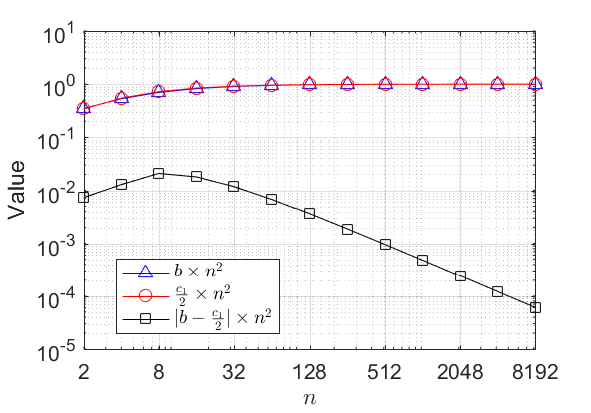}
	\caption{The differences between $b$ and $\frac{c_1}{2}$.}
	\label{fig:bs}
\end{figure}

Fig.~\ref{fig:bs} shows the values of $b$, $\frac{c_1}{2}$, and their differences on different dimensions. We use the expectation of $\|\mathbf{v}\|$ in the computation of $b$. It is obvious that $b$ and $\frac{c_1}{2}$ are quite close, and their differences are about several orders of magnitude smaller than $b$ and $\frac{c_1}{2}$. Hence, there would be only subtle performance difference between these update schemes.

\subsubsection{Cumulative Step Size Adaptation}
Apart from the distribution mean and mutation matrix, the step size is adapted independently following the cumulative step size adaptation~\cite{Hansen01}\cite{Arnold04}. Another $\mathbf{s}$-path is constructed to exploit the correlations between consecutive search directions over generations
\begin{equation}\label{spath}
\mathbf{s}_{t+1}   = (1-c_{\sigma}) \mathbf{s}_{t} +\sqrt{c_{\sigma}(2-c_{\sigma})}\sqrt{\mu_{\text{eff}}} \mathbf{z}_w.
\end{equation}
Then the step size is adapted by comparing the length of the evolution path with its expectation 
\begin{equation}
\sigma_{t+1} = \sigma_t \exp \left(\frac{c_{\sigma}}{d_{\sigma}}\left(\frac{\|\mathbf{s}_{t+1}\|}{E[\| \mathcal{N}(\mathbf{0},\mathbf{I})\|]}-1 \right) \right),
\end{equation}
where $d_{\sigma} \geq 1$ is the damping parameter. The $E[\| \mathcal{N}(\mathbf{0},\mathbf{I})\|]$ is the expected length of random vectors drawn from the standard Gaussian distribution. If consecutive directions $\mathbf{z}_w$ are correlated, the $\mathbf{s}$-path would be very long, and the step size tends to increase. If they are anti-related, the $\mathbf{s}$-path would be rather short, and the step size tends to decease.

The $\mathbf{s}$-path is an alternative to the evolution path in CMA-ES~\cite{Hansen01} that accumulates successive search directions without scaling in the local coordinate system expanded by the eigenvectors. In the proposed algorithm, it is computationally expensive to obtain the orthogonal matrix, and the $\mathbf{s}$-path is used as an alternative. 

\subsection{Algorithm Framework}
The proposed algorithm is presented in the following Algorithm~\ref{AEP}. The parameters are initialized in line 2. At each generation, a number of $\lambda$ solutions are generated in line 5-7. These new solutions are evaluated and sorted in line 9. The distribution mean and search directions are computed in line 10. Two evolution paths are constructed in line 11-12, and the matrix $\mathbf{A}$ is updated in line 13. Then the step size is adapted in line 14-15. 

\begin{algorithm}[!htb]
	\caption{The $(\mu/\mu_w,\lambda)$-MMA-ES}
	\label{AEP}
	\begin{algorithmic}[1]
		\STATE \textbf{Given:} $\lambda, \mu, w_i,  c_1, c, c_{\sigma}, d_{\sigma}$
		\STATE \textbf{Initialize:} $\mathbf{m}_{0}, \sigma_{0}, \mathbf{p}_{0} = \mathbf{0},  \mathbf{v}_{0} = \mathbf{0}, \mathbf{s}_{0} = \mathbf{0}, t=0 $
		\REPEAT
		\FOR{1 to $\lambda$}
		\STATE $\displaystyle \mathbf{z}_{i} \sim \mathcal{N}(\mathbf{0}, \mathbf{I})$
		\STATE $\displaystyle \mathbf{y}_{i} = \mathbf{A}_t \mathbf{z}_{i}  \sim \mathcal{N}(\mathbf{0}, \mathbf{A}_t \mathbf{A}_t^{T})$
		\STATE $\displaystyle \mathbf{x}_{i} = \mathbf{m}_{t} + \sigma_{t} \mathbf{y}_{i} $
		\ENDFOR
		\STATE sort $f(\mathbf{x}_{1:\lambda})\leq  \dots \leq f(\mathbf{x}_{\lambda:\lambda})$
		\STATE $\displaystyle \mathbf{m}_{t+1}  = \sum_{i=1}^{\mu}w_i\mathbf{x}_{i:\lambda}, ~\mathbf{y}_{w}  = \sum_{i=1}^{\mu}w_i\mathbf{y}_{i:\lambda}, ~\mathbf{z}_{w}  = \sum_{i=1}^{\mu}w_i\mathbf{z}_{i:\lambda}$
		\STATE $\displaystyle  \mathbf{p}_{t+1} = (1-c)\mathbf{p}_{t} + \sqrt{c(2-c)} \sqrt{\mu_{\text{eff}}}\mathbf{y}_w$
		\STATE $\displaystyle \mathbf{v}_{t+1} = (1-c)\mathbf{v}_{t} + \sqrt{c(2-c) } \sqrt{\mu_{\text{eff}}}\mathbf{z}_w$
		\STATE $\displaystyle \mathbf{A}_{t+1} = (1-\frac{c_1}{2})\mathbf{A}_{t} + \frac{c_1}{2}  \mathbf{p}_{t+1} \mathbf{v}_{t+1}^T$
		\STATE $\displaystyle \mathbf{s}_{t+1} = (1-c_{\sigma})\mathbf{s}_{t} + \sqrt{c_{\sigma}(2-c_{\sigma}) }  \sqrt{\mu_{\text{eff}}}\mathbf{z}_{w} $
		\STATE $\displaystyle \sigma_{t+1} = \sigma_{t} \cdot \exp\left(\frac{c_{\sigma}}{d_{\sigma}}\left( \frac{\| \mathbf{s}_{t+1}  \|}{E[\| \mathcal{N}(\mathbf{0},\mathbf{I})\|]}-1\right)  \right)$\\
		\STATE $t = t+1$
		\UNTIL stopping criterion is met \\
		\RETURN
	\end{algorithmic}
\end{algorithm}

The update equation for the mutation matrix $\mathbf{A}$ involves only the outer product of the evolution paths. Hence, the computational complexity is of $O(n^2)$. As the sampling procedure involves matrix-vector multi-plication, it cannot be further reduced. 

\subsection{Parameter Settings}
The involved algorithm parameters are borrowed from~\cite{Suttorp09}, and listed in Table~\ref{parameters}. The changing rate $c, c_1$ are designed to be inverse proportional to the number of free parameters in the evolution paths and the matrix. Therefore, the number of generations involved in the cumulations of $\mathbf{A}$ is $c_1^{-1} \propto n^2$. The changing rate $c_{\sigma}$ is designed to $c_{\sigma} \propto 1/\sqrt{n}$ for rapid adaptation. The setting of $c_{\sigma} \propto 1/\sqrt{n}$ or $c_{\sigma} \propto 1/n$ does not affect the algorithm performance dramatically~\cite{Beyer16}. 
\begin{table}[!htb]
	\centering 
	\caption{Algorithm parameters}
	\label{parameters}
	\begin{tabular}{l l}
		\hline 
		Population\\
		$\displaystyle \lambda = 4+ \lfloor 3\ln n\rfloor,~\mu = \lfloor\frac{\lambda}{2}\rfloor$\\
		Selection and recombination\\
		$\displaystyle w_i = \frac{\ln(\mu+1) - \ln i}{\mu\ln(\mu+1)-\sum_{j=1}^{\mu}\ln j }, ~i=1,\dots, \mu$ 	 \\
		$\mu_{\text{eff}} = \frac{1}{\sum_{i=1}^{\mu}w_i^2}$ \\ 
		step size adaptation\\
		$\displaystyle c_{\sigma} =\frac{\sqrt{\mu_{\text{eff}}}}{\sqrt{n}+\sqrt{\mu_{\text{eff}}} }, d_{\sigma } = 1+ 2\cdot \max\left(0, \sqrt{\frac{\mu_{\text{eff}} -1}{n+1}}-1 \right) +c_{\sigma} $\\
		Changing rates\\
		$\displaystyle c = \frac{4}{n+4}, c_{1} = \frac{2}{(n+\sqrt{2})^2} $ \\
		\hline
	\end{tabular}
\end{table}

\subsection{Analysis on the Evolution Path}
We analyze the orthogonality of consecutive $\mathbf{z}$-vectors of the $\mathbf{v}$-path under some reasonable assumptions. For convenience, we denote the $\sqrt{\mu_{\text{eff}}}\mathbf{z}_w$ at generation $t$ by $\mathbf{z}_t$. As previously demonstrated, the $\mathbf{v}$-path is distributed according to the standard Gaussian distribution under random selection. Hence, we make the assumptions that $\| \mathbf{v}_t\| $ and $\| \mathbf{z}_t\| $ are unchanged by selection over generation. That is, only the directions of $\mathbf{v}_{t}$ and $\mathbf{z}_{t}$ change while their lengths are kept constant. By substituting~\eqref{vpath} into $\|\mathbf{v}_{t+1}\| \approx \|\mathbf{v}_{t}\|$, we can get $\mathbf{v}_{t}^T \mathbf{z}_{t+1} \approx 0$. Given $(c_1 + c_{\mu}) \ll 1$, we have $\mathbf{A}_{t}\approx \mathbf{A}_{t+1}$. By further assuming $\mathbf{v}_{t-1}^T \mathbf{z}_{t+1} \approx 0$, we can derive that
\begin{equation}
\mathbf{z}_{t}^T \mathbf{z}_{t+1} \approx 0.
\end{equation}
This means that consecutive mutation steps $\mathbf{z}_{t}$ and $\mathbf{z}_{t+1}$ tends to be orthogonal. This can be equivalently written as
\begin{equation}
\left(\mathbf{m}_t - \mathbf{m}_{t-1}\right)^T \mathbf{C}_t^{-1}\left(\mathbf{m}_{t+1} - \mathbf{m}_{t}\right) \approx 0,
\end{equation}
where $\mathbf{C}_t = \mathbf{A}_t\mathbf{A}_t^T$. This indicates that successive search directions are conjugate in terms of the covariance matrix. As the covariance matrix generally converges to the inverse Hessian matrix on quadratic objective functions (up to a scalar factor)~\cite{Hansen01}\cite{Beyer14}, this is desirable for efficient search. 

\section{Experimental Studies}

We investigate the algorithm behaviors and performances on a set of commonly used test problems defined in table~\ref{tableFuncDef}. We focus on the convergence properties of the algorithm. The initial distribution mean $\mathbf{m}_0$ is uniformly randomly sampled in $[-10, 10]^n$, and the step size $\sigma_0$ is initialized to be $20/3$. The algorithm runs 21 times independently on each test problem. In the comparison of the experimental results, we conduct Wilcoxon rank sum test to assess the difference. When we state that two results are different, we have assert its statistical significance at the 5\% significance level.
\begin{table}[!htb]
	\centering 
	\caption{Test Problems}
	\label{tableFuncDef}
	\begin{tabular}{l l}
		\hline \hline
		Function		& Target\\
		\hline
		$  f_{\text{Sp}}(\mathbf{x})= \sum_{i=1}^n x_i^2$ 	& $10^{-10}$ \\
		$  f_{\text{Cig}}(\mathbf{x})= x_1^2+10^6\cdot\sum_{i=2}^n x_i^2$	& $10^{-10}$ \\
		$  f_{\text{Ctb}}(\mathbf{x})=  x_1^2+10^4\cdot\sum_{i=2}^{n-1} x_i^2 +10^6 \cdot x_n^2$	& $10^{-10}$ \\
		$  f_{\text{Ell}}(\mathbf{x})= \sum_{i=1}^n 10^{6\cdot\frac{i-1}{n-1}}\cdot x_i^2$ 	& $10^{-10}$ \\
		$  f_{\text{Tab}}(\mathbf{x})=  10^6\cdot x_1^2+\sum_{i=2}^n x_i^2$	& $10^{-10}$ \\
		$  f_{\text{Tx}}(\mathbf{x})=  \sum_{i=1}^{\lfloor n/2\rfloor}x_i^2+10^6\cdot \sum_{i=\lfloor n/2\rfloor+1}^n x_i^2$	& $10^{-10}$ \\
		$  f_{\text{dP}}(\mathbf{x})=  \sum_{i=1}^{n} x_{i}^{2+4 (i-1)/(n-1) }$	& $10^{-10}$ \\
		$  f_{\text{Sch}}(\mathbf{x})=  \sum_{i=1}^{n} (\sum_{j=1}^{i-1}x_j)^2$	& $10^{-10}$ \\
		$  f_{\text{Ros}}(\mathbf{x})=  \sum_{i=1}^{n-1}(100(x_{i+1} - x_i^2)^2 + (x_i-1)^2)$	& $10^{-10}$ \\
		$  f_{\text{pR}}(\mathbf{x})=  -x_1 + 100\sum_{i=2}^{n}x_i^2$			& $-10^{10}$ \\
		\hline 
	\end{tabular}
\end{table}

\subsection{Experiments on the Approximations}
We experimentally validate the effectiveness of the approximation of the $\mathbf{v}$-path to the inverse vector of the $\mathbf{p}$-path, and the approximation of coefficients. 

\subsubsection{Approximation to the Inverse Vector}
We first conduct experiments to validate the approximation of the $\mathbf{v}$-path to the inverse vector $\mathbf{u}$. As the $\mathbf{v}$-path is distributed according to the standard Gaussian distribution under the stationarity condition, we care only the direction of $\mathbf{v}$. At each generation, we compute the $\mathbf{v}_{t+1}$ using~\eqref{vpath} and the inverse vector $\mathbf{u}_{t+1} = \mathbf{A}_t^{-1}\mathbf{p}_{t+1}$. The similarity between $\mathbf{u}_{t+1}$ and $\mathbf{v}_{t+1}$ is measured by 
\begin{equation}
\alpha= 1- \frac{\mathbf{v}_{t+1}^T\mathbf{u}_{t+1}}{\|\mathbf{v}_{t+1}\|\cdot \|\mathbf{u}_{t+1}\|}.
\end{equation}
It indicates that $\mathbf{v}$ is close to $\mathbf{u}$ if $\alpha\approx 0 $. The experiments are conducted on $n=32$. 

\begin{figure}[!htb]
	\centering
	\subfigure[$f_{\text{Cig}}$]{
		\includegraphics[width=0.45\linewidth]{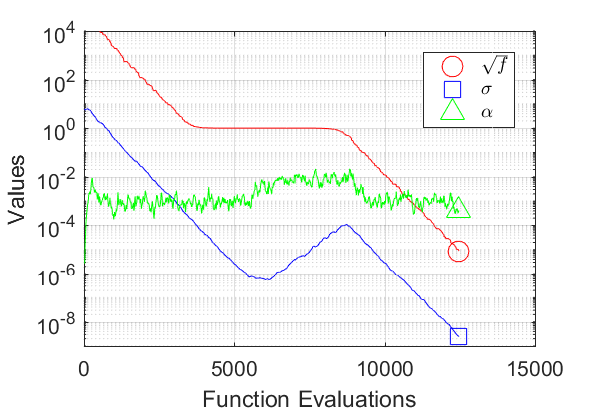}}
	\subfigure[$f_{\text{Ell}}$]{
		\includegraphics[width=0.45\linewidth]{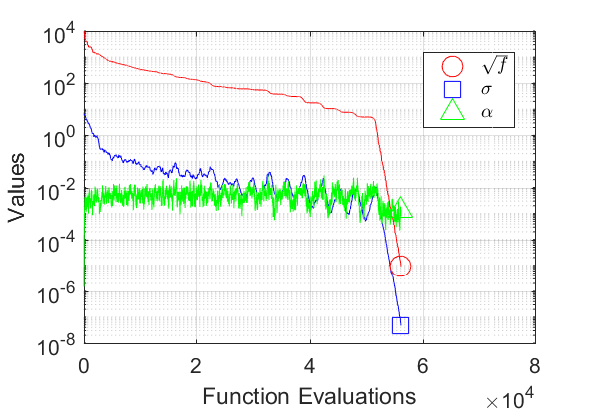}}   
	\subfigure[$f_{\text{Tab}}$]{
		\includegraphics[width=0.45\linewidth]{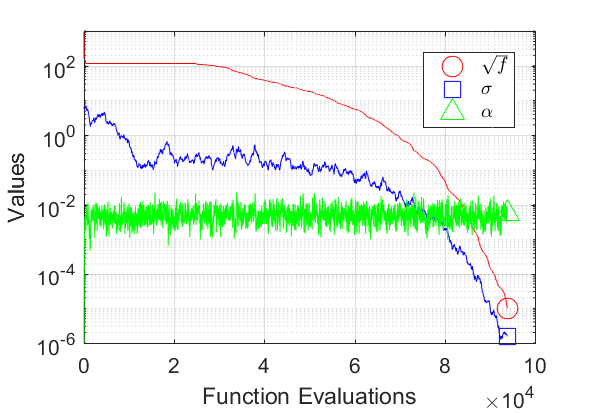}} 
	\subfigure[$f_{\text{Ros}}$]{
		\includegraphics[width=0.45\linewidth]{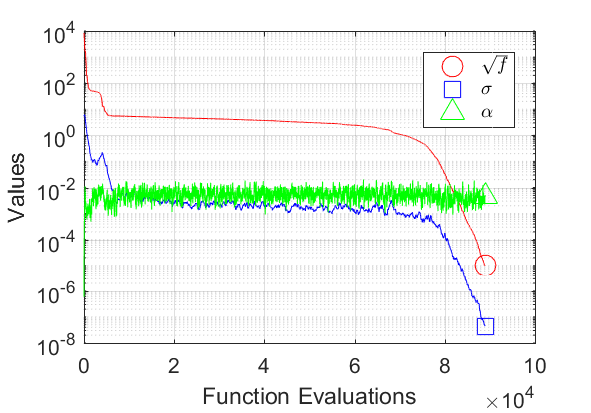}}
	\caption{The approximation of the $\mathbf{v}$-path to the inverse vector. The experiments are conducted on dimension 32, and presented are the median run out of 21 independent runs.}
	\label{EPfits}
\end{figure}

Fig.~\ref{EPfits} presents the experimental results on some test problems. It shows that $\alpha$ is of the order $10^{-3}$ to $10^{-2}$ in the whole optimization procedure on the test problems, including the $f_{\text{Ros}}$ function which has a parabolic shaped valley. Consequently, the $\mathbf{v}$-path approximates well to the inverse vector $\mathbf{u}$. 

\begin{figure}[!htb]
	\centering
	\includegraphics[width=0.75\linewidth]{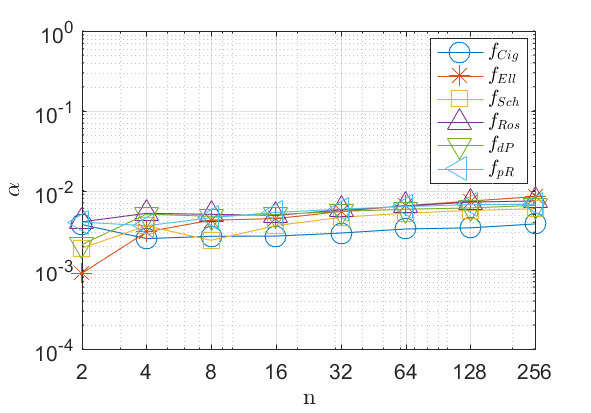}
	\caption{The similarity $\alpha$ on different dimensions, averaged over 21 runs.}
	\label{aexpall}
\end{figure}

We investigate the $\alpha$ on different dimensions. On each dimension, it is averaged over 21 independent runs. The experimental results are shown in Fig.~\ref{aexpall}. Clearly, $\alpha$ is of the order $10^{-3}$ to $10^{-2}$ on all dimensions of the test problems, including on $f_{\text{Ros}}$ which has a parabolic shaped valley. Hence, the $\mathbf{v}$-path can be used as a good approximation to the inverse vector.

\subsubsection{Approximation of the Coefficient}
\begin{figure}[!htb]
	\centering
	\subfigure[$f_{\text{Cig}}$]{
		\includegraphics[width=0.45\linewidth]{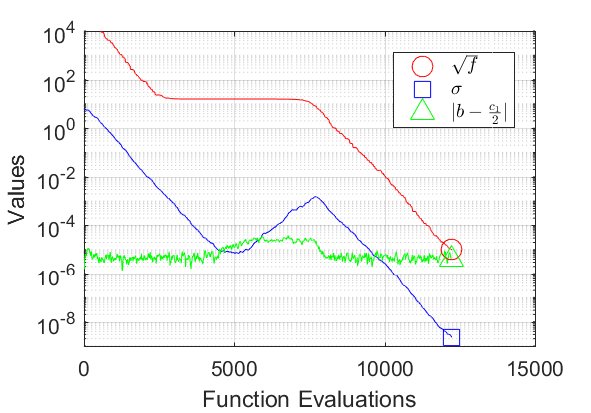}} \vspace{0pt}
	\subfigure[$f_{\text{Ell}}$]{
		\includegraphics[width=0.45\linewidth]{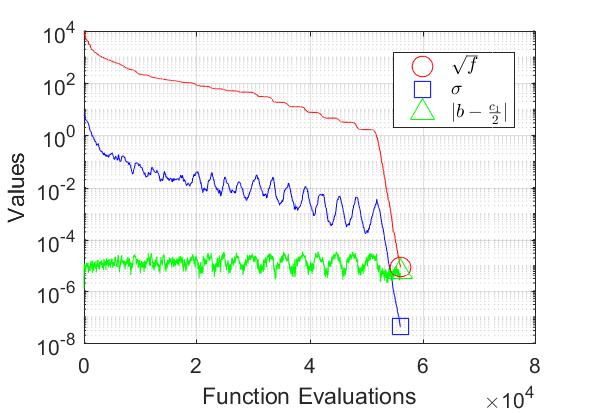}}   
	\subfigure[$f_{\text{Tab}}$]{
		\includegraphics[width=0.45\linewidth]{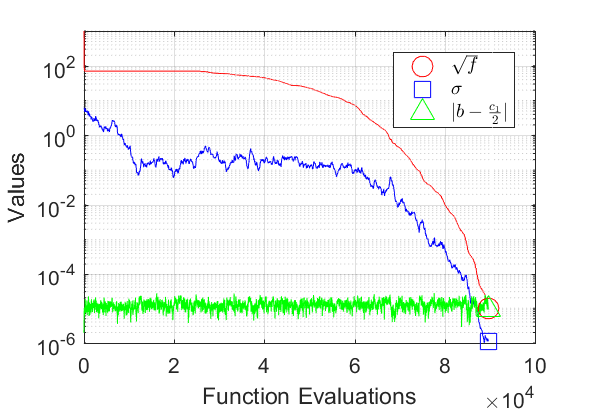}} 
	\subfigure[$f_{\text{Ros}}$]{
		\includegraphics[width=0.45\linewidth]{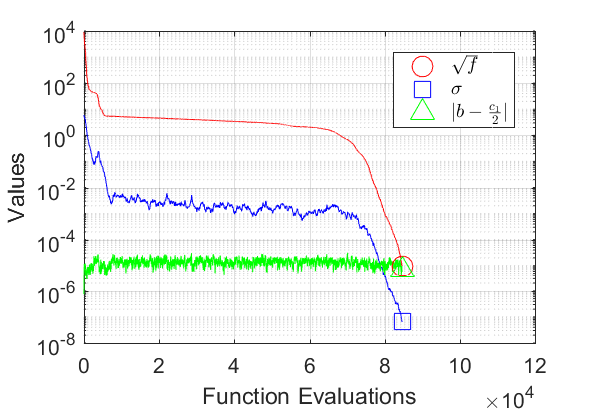}}
	\caption{The difference $|b-\frac{c_1}{2}|$ in the optimization procedure. The experiments are conducted on dimension 32, and presented are the median run out of 21 independent runs.}
	\label{HESb}
\end{figure}

We then investigate the difference between $|b-\frac{c_1}{2}|$ in MMA-ES with $b$ given by~\eqref{bcoeff}. At each generation, we compute $| b-\frac{c_1}{2}| $. Fig.~\ref{HESb} presents the experimental results on $n=32$. It clearly shows that the difference $| b-\frac{c_1}{2}| $ is about the order of $10^{-5}$ in the whole optimization procedure, indicating that the rank one update~\eqref{HES} is a good approximation to the Cholesky update. 

\begin{figure}[!htb]
	\centering
	\includegraphics[width=0.75\linewidth]{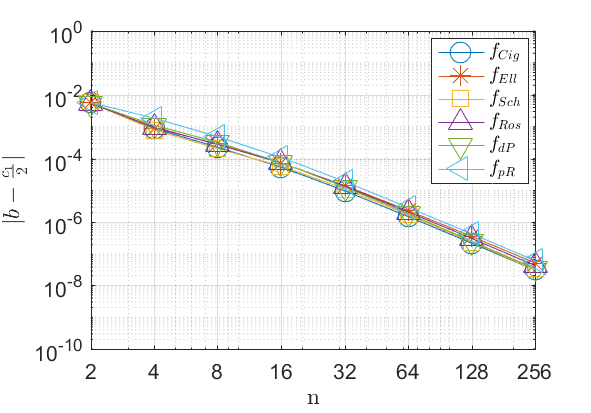}
	\caption{The difference $|b-\frac{c_1}{2}|$ on different dimensions, averaged over 21 runs.}
	\label{aepall}
\end{figure}

We further investigate the difference $|b-\frac{c_1}{2}|$ on different dimensions, which is averaged over the whole optimization procedure of 21 runs on each dimension. The experimental results in Fig.~\ref{aepall} show that $|b-\frac{c_1}{2}|$ descends from $10^{-2}$ to $10^{-8}$ as the dimension increases from of $n=2$ to $n=256$. Hence, the MMA-ES provides a good approximation to the Cholesky update.  

\subsection{Algorithm Behaviors}
We investigate the algorithm behaviors in this subsection, including the evolution of the eigenvalues of the covariance matrix, and the performance invariance under rotation in the search space. 

\subsubsection{Evolution of the Eigenvalues}
\begin{figure}[!htb]
	\centering
	\includegraphics[width=0.75\linewidth]{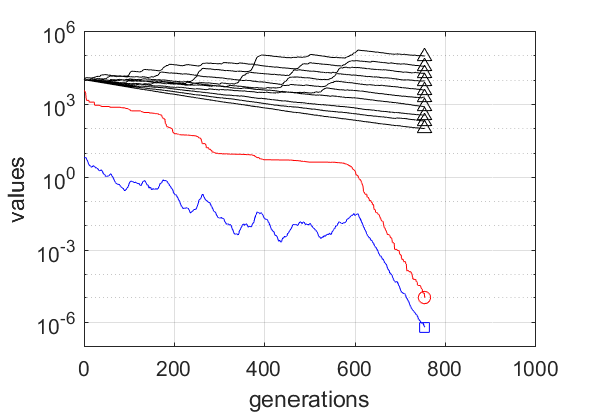}
	\caption{The evolution of the eigenvalues of $\mathbf{C}=\mathbf{AA}^T$. Shown are the objective values (red), the step size (blue), and the eigenvalues (black) of the whole optimization procedure.}
	\label{eigs}
\end{figure}

We investigate the evolution of the eigenvalues of $\mathbf{C} = \mathbf{AA}^T$ in the optimization procedure. At each generation, we decompose to $\mathbf{C}_t = \mathbf{B}\mathbf{D}^2\mathbf{B}^T$, where $\mathbf{D} = diag(d_{1},\dots, d_n)$ with $d_i$ denoting the square root of the $i$-th eigenvalue of $\mathbf{C}_t$. $d_i$ represents the scale of the variables along each eigenvector. Fig.~\ref{eigs} shows the experiential result on $f_{\text{Ell}}$ with $n=10$. The algorithm gradually learns the relative scales along the eigenvectors in the optimization procedure, and once the mutation matrix learns the relative scales, the objective value and step size decrease rapidly.  

\subsubsection{Invariance Under Rotation in the Search Space}
The main advantage of CMA-ES is its invariance properties under general linear transformation in the search space, especially the rotation. We investigate the invariance under the rotation of the search space. We generate a orthogonal matrix $\mathbf{R}$ via the Gram–Schmidt process, and investigate the performance of the algorithm on $f(\mathbf{Rx})$. 

\begin{figure}[!htb] 
	\centering
	\subfigure[$f_{\text{Ell}}$]{
		\includegraphics[width=0.45\linewidth]{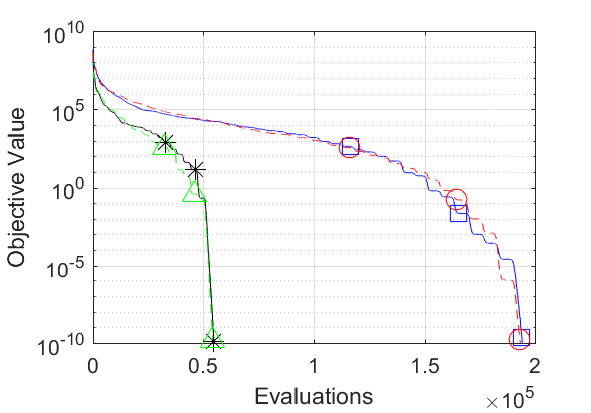}}
	\subfigure[$f_{\text{Ros}}$]{
		\includegraphics[width=0.45\linewidth]{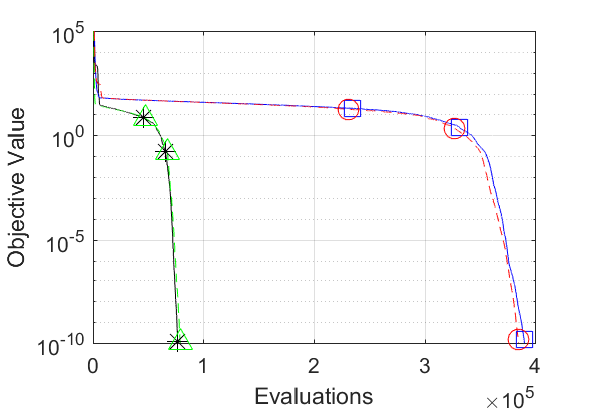}}
	\caption{The objective trajectories on separable and non-separable objective functions with $n=32 $ and $n=64$. Shown are the median run out of 21 independent runs for each objective function. }
	\label{Rotfigs}
\end{figure}

Fig.~\ref{Rotfigs} shows the median run out of 21 independent runs on $f_{\text{Ell}}$ and $f_{\text{Ros}}$ with both $n=32 $ and $n=64$. It clearly shows that the algorithm performs uniformly on the $f(\mathbf{x})$ and $f(\mathbf{Rx})$. The slight differences are mainly due to the initialization of the distribution mean $\mathbf{m}_{0}$. Hence, the algorithm performance remains invariant under rotation in the search space. 

\subsection{Studies on Algorithm Performances}
\subsubsection{Algorithm Performances Comparison}
we compare the algorithms performances on the test problems. We present the median run in terms of the number of function evaluations of each algorithm. As the $f_{\text{Ros}}$ function has a local minimum~\cite{Rosenbrock} and the algorithms may fail, we consider that the failed runs cost more objective function evaluations than any successful runs.

\begin{figure*}[!htb] 
	\centering
	\subfigure[$f_{\text{Sp}}$]{
		\includegraphics[width=0.32\linewidth]{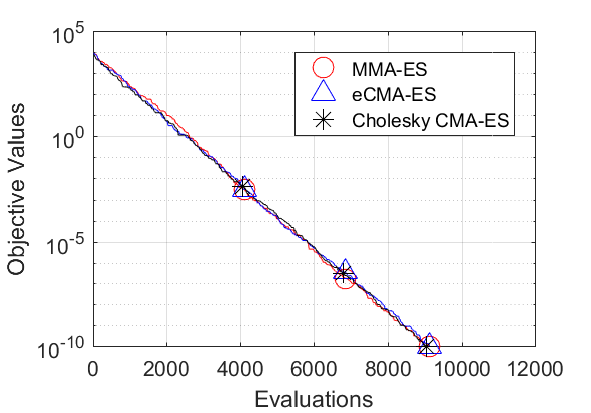}} 
	\subfigure[$f_{\text{Cig}}$]{
		\includegraphics[width=0.32\linewidth]{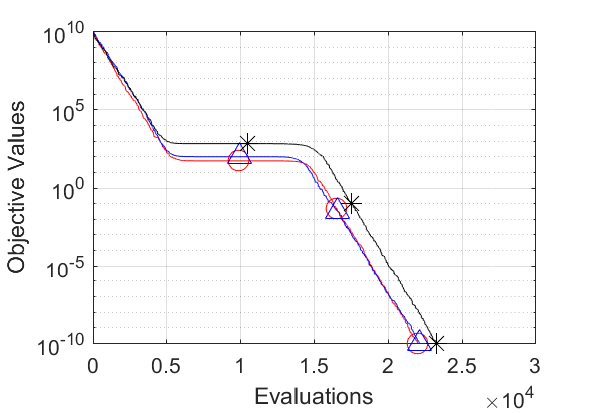}}
	\subfigure[$f_{\text{Ell}}$]{
		\includegraphics[width=0.32\linewidth]{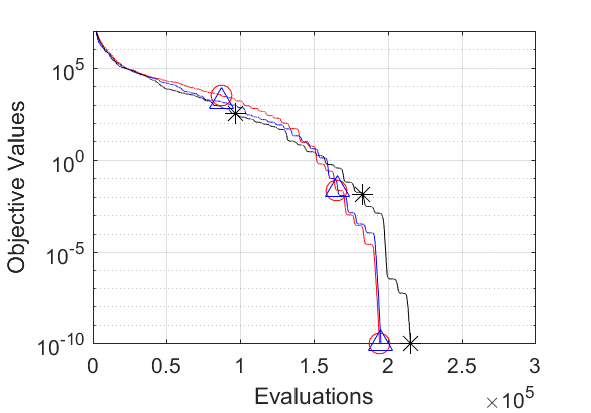}}  
	\subfigure[$f_{\text{Tab}}$]{
		\includegraphics[width=0.32\linewidth]{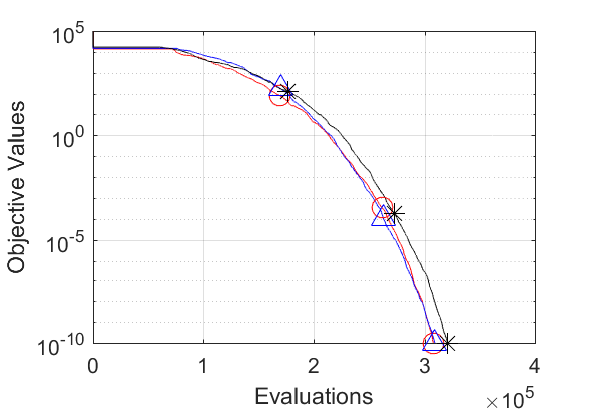}} 
	\subfigure[$f_{\text{Tx}}$]{
		\includegraphics[width=0.32 \linewidth]{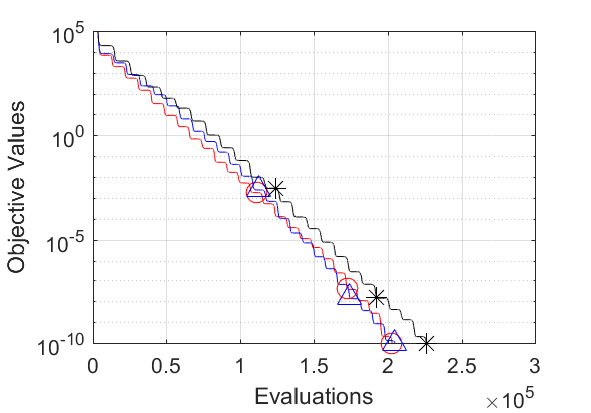}}   
	\subfigure[$f_{\text{dP}}$]{
		\includegraphics[width=0.32\linewidth]{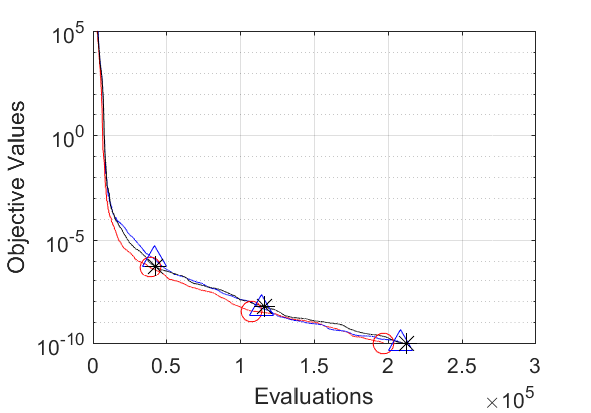}} 
	\subfigure[$f_{\text{Ros}}$]{
		\includegraphics[width=0.32\linewidth]{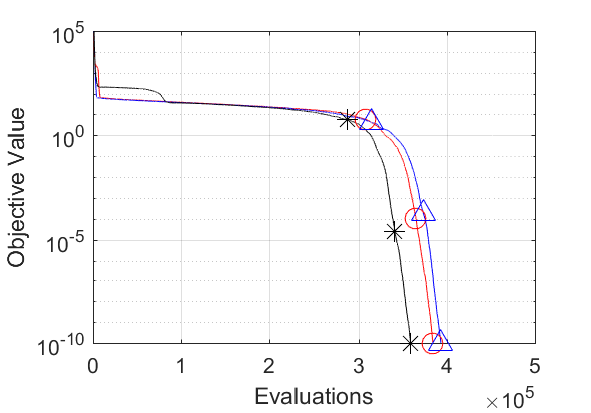}} 
	\subfigure[$f_{\text{Sch}}$]{
		\includegraphics[width=0.32\linewidth]{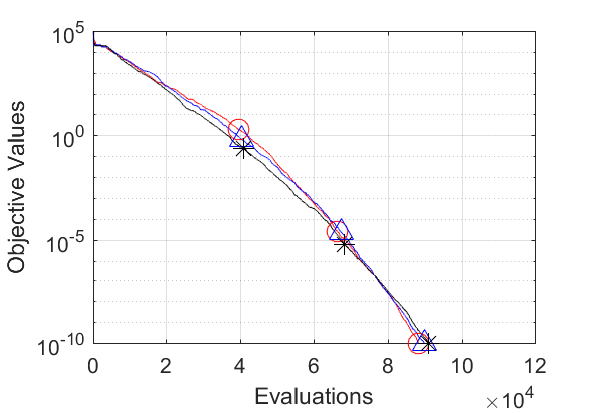}} 
	\subfigure[$f_{\text{pR}}$]{
		\includegraphics[width=0.32\linewidth]{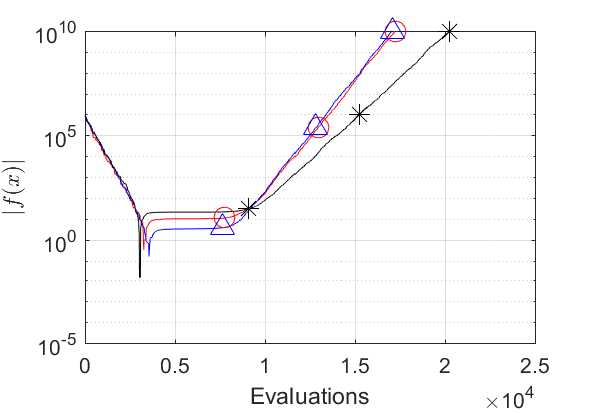}}
	\caption{The objective trajectories of each algorithm in dimension 64. Shown are the median run out of 21 independent runs.}
	\label{AEPfits}
\end{figure*}

Fig. \ref{AEPfits} shows the experimental results. The most important observation is that MMA-ES performs quite well on all the test problems. It outperforms Cholesky CMA-ES on most of the test problems with ill-conditioned Hessian matrix, and the $f_{\text{pR}}$ with a parabolic ridge. This validates the effectiveness of MMA-ES update. On the $f_{\text{Sp}}$, MMA-ES performs similarly to Cholesky CMA-ES. On the $f_{\text{Ros}}$, MMA-ES costs slightly more function evaluations than that of Cholesky CMA-ES. This may due to the parabolic shaped valley in the fitness landscape of $f_{\text{Ros}}$, which requires rapid adaptation of the evolution path and the $\mathbf{v}$-path may lay behind. MMA-ES performs quite similar on the problems, while the update scheme is much simpler.

\subsubsection{Scaling on Dimension}

We investigate the algorithm scalability on dimension of the search space. The algorithm scalability is measured by the average number of objective function evaluations (FE) on dimension. Because the $f_{\text{Ros}}$ is multi-modal~\cite{Rosenbrock}, we use the success performance $\hat{SP} = E_s/p_s$, where $E_s$ is the average function evaluations of successful runs, and $p_s$ is the success ratio of all independent runs~\cite{Auger05}\cite{Hansen11}.

\begin{figure*}[!htb]
	\centering
	\subfigure[$f_{\text{Cig}}$]{
		\includegraphics[width=0.32\linewidth]{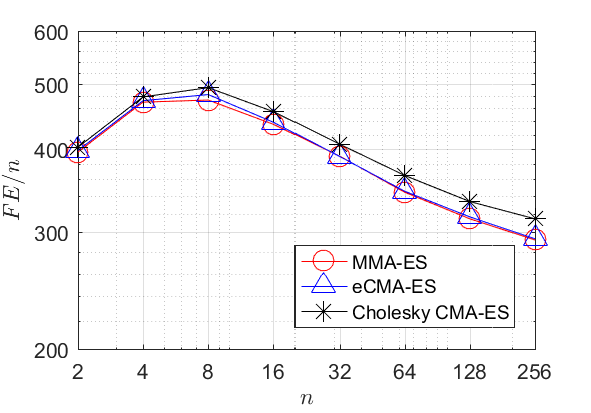}}  
	\subfigure[$f_{\text{Ctb}}$]{
		\includegraphics[width=0.32\linewidth]{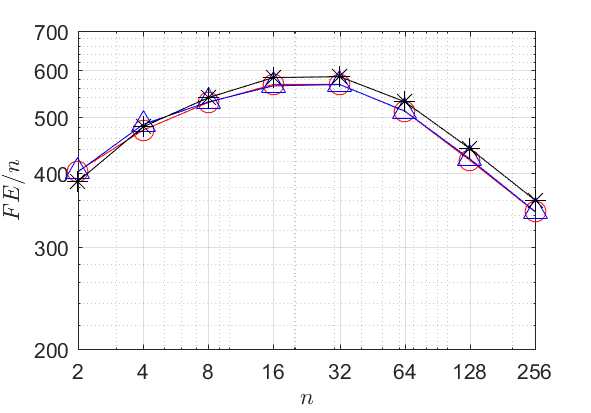}}  
	\subfigure[$f_{\text{Tab}}$]{
		\includegraphics[width=0.32\linewidth]{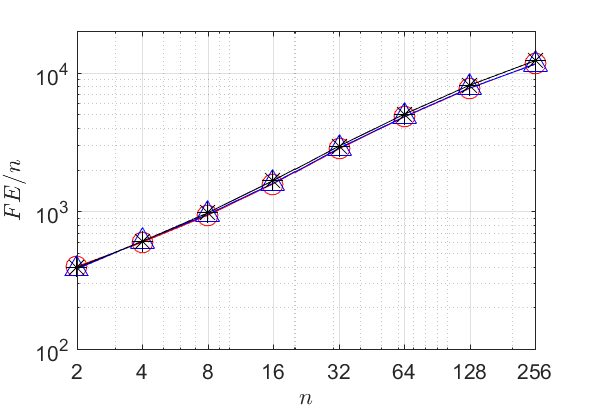}}
	\subfigure[$f_{\text{Cig}}$]{
		\includegraphics[width=0.32\linewidth]{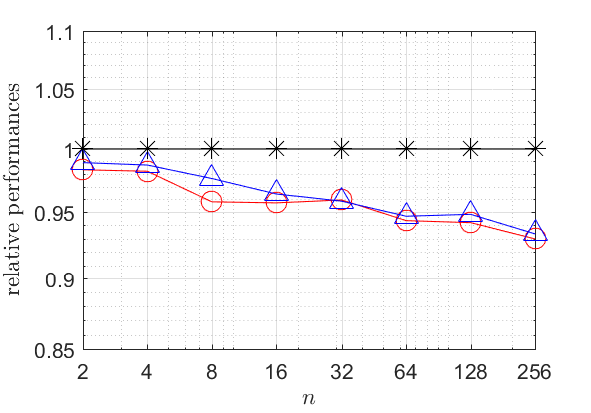}}   
	\subfigure[$f_{\text{Ctb}}$]{
		\includegraphics[width=0.32\linewidth]{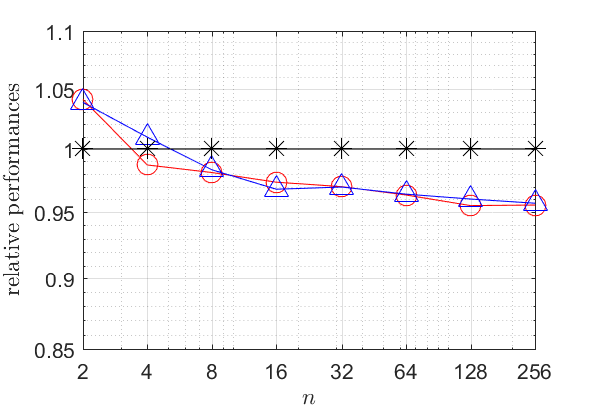}} 
	\subfigure[$f_{\text{Tab}}$]{
		\includegraphics[width=0.32\linewidth]{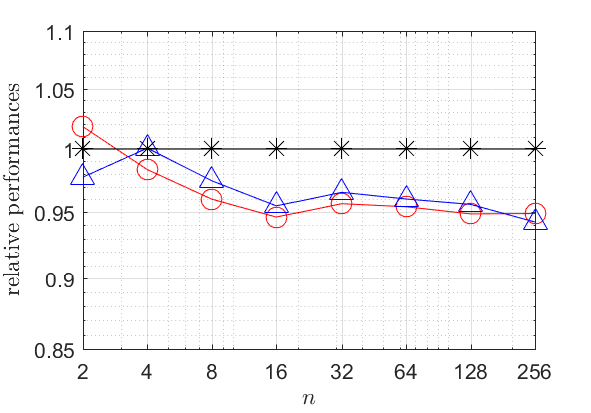}} 
	\subfigure[$f_{\text{Ell}}$]{
		\includegraphics[width=0.32\linewidth]{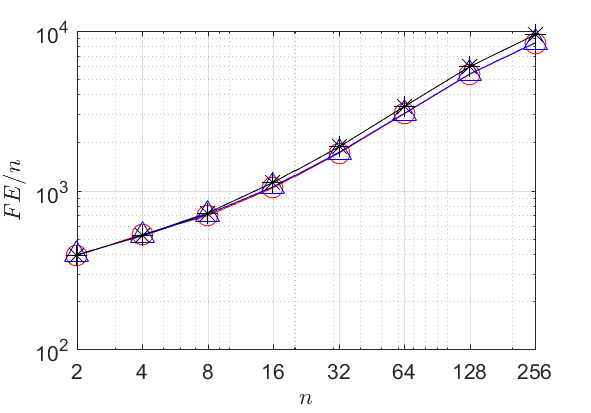}} 
	\subfigure[$f_{\text{Tx}}$]{
		\includegraphics[width=0.32\linewidth]{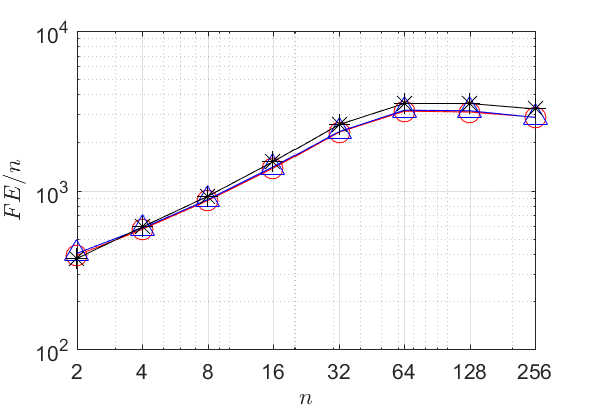}}
	\subfigure[$f_{\text{dP}}$]{
		\includegraphics[width=0.32\linewidth]{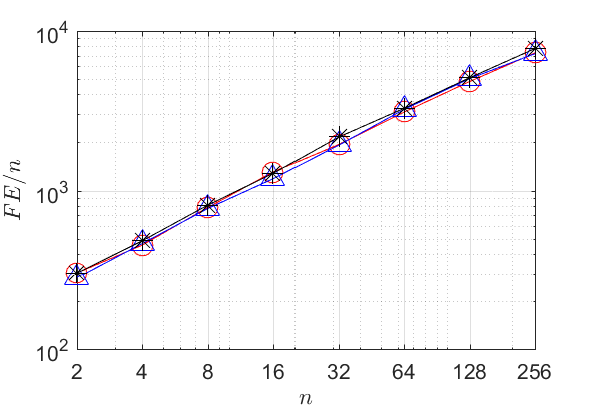}}
	 \subfigure[$f_{\text{Ell}}$]{
	 	\includegraphics[width=0.32\linewidth]{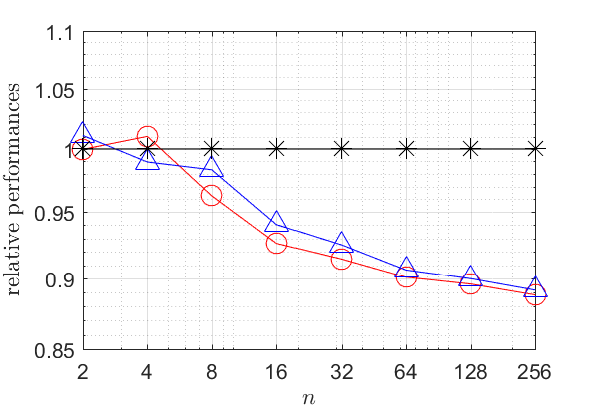}}
 	\subfigure[$f_{\text{Tx}}$]{
 		\includegraphics[width=0.32\linewidth]{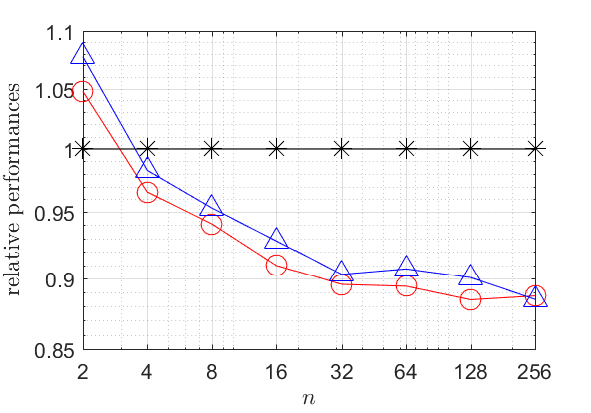}} 
 		\subfigure[$f_{\text{dP}}$]{
 		\includegraphics[width=0.32\linewidth]{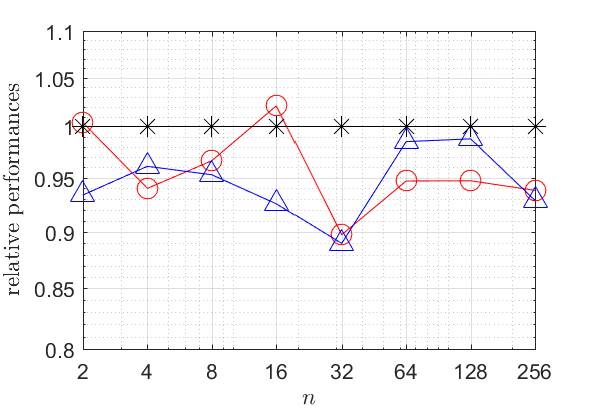}}
	\caption{The average number of function evaluations (FE) required to reach the predefined accuracy, and the relative performance of two algorithms (I). The algorithms run 21 times independently on each dimension of the problems.}
	\label{Eval_q1}
\end{figure*}

\begin{figure*}[!htb]
	\centering
	\subfigure[$f_{\text{Ros}}$]{
		\includegraphics[width=0.32\linewidth]{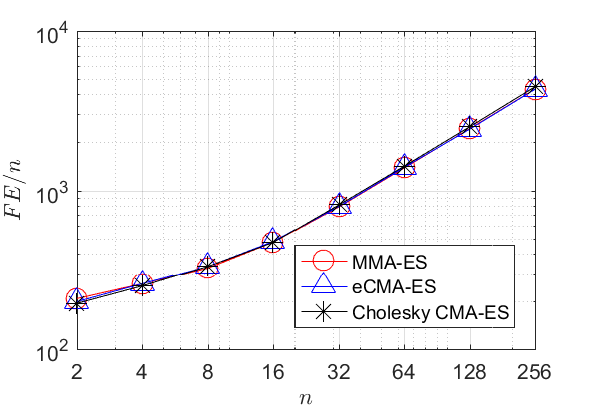}} 
	\subfigure[$f_{\text{Sch}}$]{
		\includegraphics[width=0.32\linewidth]{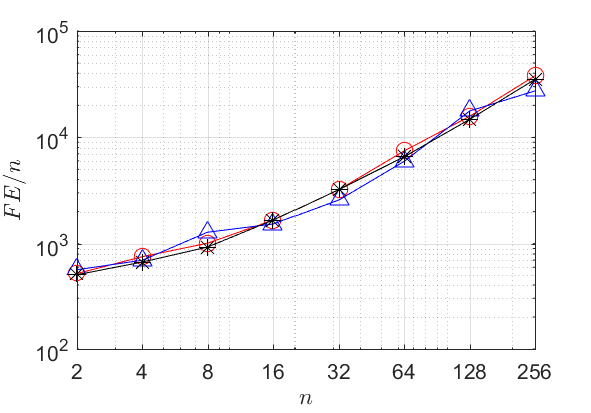}} 
	\subfigure[$f_{\text{pR}}$]{
		\includegraphics[width=0.32\linewidth]{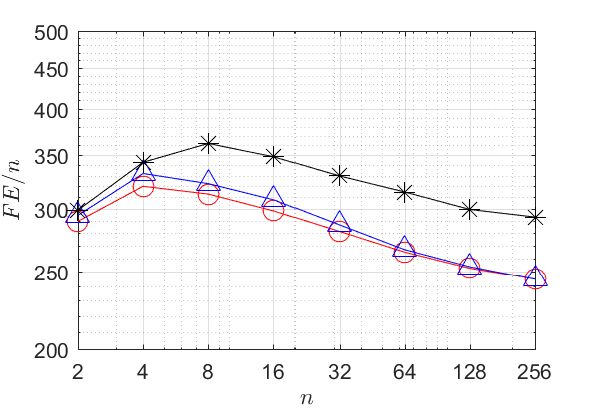}}  
	\subfigure[$f_{\text{Ros}}$]{
		\includegraphics[width=0.32\linewidth]{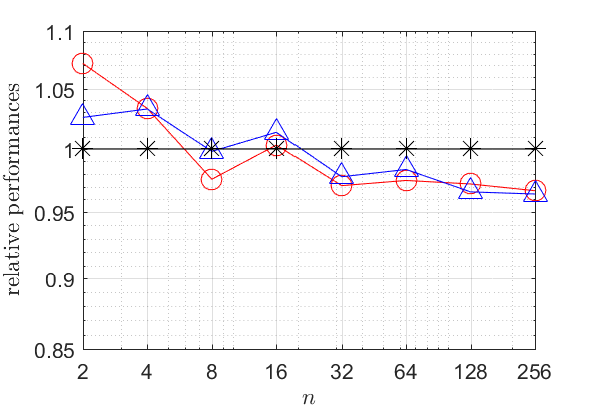}} 
	\subfigure[$f_{\text{Sch}}$]{
		\includegraphics[width=0.32\linewidth]{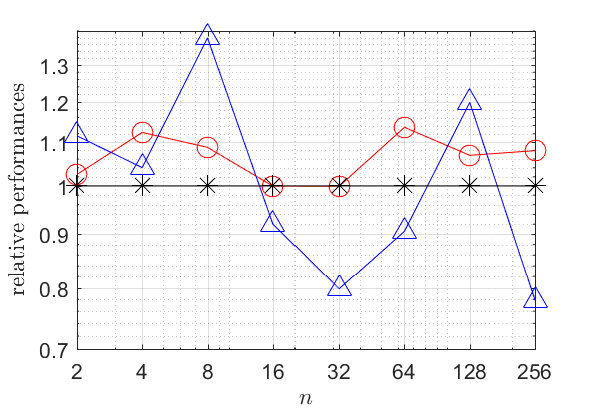}}
	\subfigure[$f_{\text{pR}}$]{
		\includegraphics[width=0.32\linewidth]{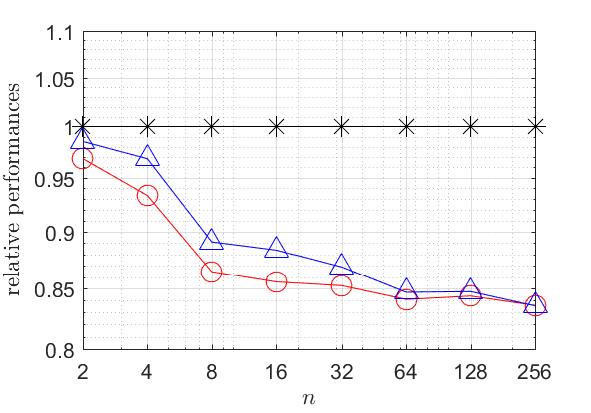}}
	\caption{The average number of function evaluations (FE) required to reach the predefined target, and the relative performance of two algorithms (II). The algorithms run 21 times independently on each dimension of the problems.}
	\label{Eval_q2}
\end{figure*}

The experimental results in Fig.~\ref{Eval_q1} and Fig.~\ref{Eval_q2} show that the proposed method scales similarly to the Cholesky CMA-ES, while presenting some advantages on ill-conditioned problems. This indicates that, with increasing problem dimension, the proposed simplified update scheme works well. 

To make the difference more clear, we investigate the relative performance of the proposed algorithms, with the Cholesky CMA-ES as baseline. The relative performance is measured by 
\begin{equation}
\beta = \frac{FE_2}{FE_{1}},
\end{equation}
where $FE_{1}$ denotes the average number of function evaluations of the Cholesky CMA-ES, and $FE_{2}$ denotes that of the MMA-ES and the eCMA-ES. It measures how much times of function evaluations are required for the compared algorithms to reach the predefined accuracy. If $\frac{FE_2}{FE_{1}}<1$, the compared algorithm costs less function evaluations than Cholesky CMA-ES. The experimental results show that $\frac{FE_2}{FE_{1}}<1$ on most of the test problems and dimensions. Actually, Cholesky CMA-ES only outperforms MMA-ES and eCMA-ES on some dimensions on $f_{\text{Ros}}$ and $f_{\text{Sch}}$.

\section{Conclusion}
In this paper, we have proposed an efficient rank one update for Cholesky CMA-ES by using a new $\mathbf{v}$-path. It is as simple as the rank one update of CMA-ES, while avoiding the computational matrix decomposition. The $\mathbf{v}$-path accumulates $\mathbf{z}$-vectors drawn from the standard Gaussian distribution, and approximates the inverse vector of the $\mathbf{p}$-path. 

We have analyzed the orthogonal properties between consecutive $\mathbf{z}$-vectors. We have conducted experiments to verify that the effectiveness of the approximations, and investigated the algorithm behaviors. Then we have showed that it outperforms or performs competitively to Cholesky CMA-ES. We have investigated the algorithm scalability on dimension.

The proposed method can be directly extended to the limited memory variant to handle large scale optimization problems following the idea of~\cite{Loshchilov17}\cite{Zhenhua2017b}. The sparsity technique~\cite{MeyerNiebergK15} can be used here without conducting matrix decomposition. The update~\eqref{HES} (especially for the case without cumulation in the evolution paths with $c=1$) shows some similarity to the Hebbian learning rule~\cite{Sanger89GHA}. The relation is to be further explored.


\bibliographystyle{plain}
\bibliography{bibtex2}

\section*{Appendix}

\subsection{Bound of the Approximation}
We illustrate that the approximation of the $\mathbf{v}$-path to the inverse vector is upper bounded. For convenience, we denote the weighted mean mutation $\sqrt{\mu_{\text{eff}} } \mathbf{z}_w$ at generation $t$ by $\mathbf{z}_t$. Then the update equations can be written as
\begin{equation}
\mathbf{p}_{t+1} = a \mathbf{p}_{t} + b \mathbf{A}_{t} \mathbf{z}_{t}, ~
\mathbf{v}_{t+1} = a \mathbf{v}_{t} + b \mathbf{z}_{t}, ~
\mathbf{u}_{t+1} = \mathbf{A}^{-1}_{t} \mathbf{p}_{t+1}, \label{E3}
\end{equation}
where $a = 1-c<1, b = \sqrt{c(2-c)} <1$. 

The evolution paths $\mathbf{p}_{t+1}$ and $\mathbf{v}_{t+1}$ can be rewritten as 
\begin{equation}
\mathbf{p}_{t+1} = b \sum_{i=1}^{t} a^{t-i} \mathbf{A}_{i} \mathbf{z}_{i}, ~ \mathbf{v}_{t+1} = b \sum_{i=1}^{t} a^{t-i} \mathbf{z}_{i}.
\end{equation}
Substituting $\mathbf{p}_{t+1}$ to $\mathbf{u}_{t+1}$, we can write it as
\begin{equation}
\mathbf{u}_{t+1} = b \sum_{i=1}^{t} a^{t-i} \mathbf{A}^{-1}_{t} \mathbf{A}_{i} \mathbf{z}_{i}.
\end{equation}
Hence, we have
\begin{align*}\label{uvdiff2}
\|\mathbf{v}_{t+1}-\mathbf{u}_{t+1}\| &\leq b\sum_{i=0}^{t} a^{t-i} \|(\mathbf{I} - \mathbf{A}_{t}^{-1} \mathbf{A}_i ) \mathbf{z}_i\| \\
& \leq b\sum_{i=0}^{t} a^{t-i}\|\mathbf{I} - \mathbf{A}_{t}^{-1} \mathbf{A}_i \|\| \mathbf{z}_i\|.
\end{align*}
We study the expectation of the right hand side. Since $\mathbf{z}_i \sim \mathcal{N}(\mathbf{0},\mathbf{I})$, we have $\mathbb{E}[\| \mathbf{z}_i\|] =\sqrt{n}$. We assume the non-singular sequence of $\mathbf{A}_t \to \mathbf{A}$ as $t\to \infty$ and $\|\mathbf{A}_t\| $ and $\|\mathbf{A}_t^{-1} \|$ are bounded $\|\mathbf{A}_t \| <L_1,~\|\mathbf{A}_t^{-1} \| <L_2$ and $L = \max(L_1, L_2)$. Specifically, for any given $\epsilon >0$, there is some $K$ such that $\|\mathbf{A}_{t}-\mathbf{A}_i\| < \frac{\epsilon}{L}$ for any $i>K$. Therefore, we have 
\begin{align*}
\|\mathbf{I} - \mathbf{A}_{t}^{-1} \mathbf{A}_i \| &=\|\mathbf{A}_{t}^{-1}(\mathbf{A}_{t}-\mathbf{A}_i)\|\\
&\leq \|\mathbf{A}_{t}^{-1}\| \|\mathbf{A}_{t}-\mathbf{A}_i )\| \\
&< \epsilon.
\end{align*}
Therefore, we have the expectation (in terms of $\mathbf{z}_t$)
\begin{align*}
~E \left[b\sum_{i=0}^{t} a^{t-i}\|\mathbf{I} - \mathbf{A}_{t}^{-1} \mathbf{A}_i \|\| \mathbf{z}_i\| \right] 
&= b \sqrt{n} \sum_{i=0}^{t} a^{t-i}\|\mathbf{I} - \mathbf{A}_{t}^{-1} \mathbf{A}_i \| \\
&\leq b \sqrt{n}\sum_{i=0}^{K} a^{t-i} \|\mathbf{I} - \mathbf{A}_{t}^{-1} \mathbf{A}_i \|+ \epsilon \sqrt{n}\sum_{i=K+1}^{t} a^{t-i}  \\
&< M b \sqrt{n}\sum_{i=0}^{K} a^{t-i}+ \epsilon \sqrt{n}\sum_{i=K+1}^{t} a^{t-i}  \\
&<  \frac{\sqrt{n}}{1-a} (Mb a^{t-K}+ \epsilon).
\end{align*}
where $M=\max_{1\leq i \leq K}\|\mathbf{I} - \mathbf{A}_{t}^{-1} \mathbf{A}_i \| $. Recall that $a = (1-c)<1 $, we have that $\|\mathbf{v}_{t+1}-\mathbf{u}_{t+1}\|$ is upper bounded by $\frac{\sqrt{n}b}{1-a} (M a^{t-K}+ \epsilon)$, where $a^{t-K} \to 0$ as $t \to \infty$.  

Both experimental and theoretical analysis indicate that the covariance matrix of CMA-ES approximates the inverse Hessian on quadratic functions (up to a scalar factor)~\cite{Hansen01}\cite{Beyer14}. Hence, it is reasonable to make the convergence assumption. 

\end{document}